\documentclass{article} 
\usepackage{iclr2025_conference,times}
\iclrfinalcopy

\usepackage{amsmath,amsfonts,bm}

\def\eqref#1{equation~\ref{#1}}

\def\1{\bm{1}}

\def\mX{{\bm{X}}}
\def\mY{{\bm{Y}}}

\DeclareMathAlphabet{\mathsfit}{\encodingdefault}{\sfdefault}{m}{sl}
\SetMathAlphabet{\mathsfit}{bold}{\encodingdefault}{\sfdefault}{bx}{n}

\usepackage{hyperref}
\usepackage{url}

\usepackage{multirow}
\usepackage{graphicx}
\usepackage{adjustbox}
\usepackage{algorithm}
\usepackage{algcompatible}
\usepackage{breqn}
\usepackage{amsmath}
\usepackage{subcaption}
\usepackage{wrapfig}

\usepackage{graphicx} 

\title{Pruning Deep Convolutional Neural Network Using Conditional Mutual Information}

 \author{Tien Vu-Van$^{*}$, Dat Du Thanh$^{*}$, Nguyen Ho$^{\dagger}$, Mai Vu$^{\ddagger}$ \\
 * Faculty of Computer Science and Engineering, 
 Ho Chi Minh City University of Technology, Vietnam\\
$\dagger$ Computer Science Department, Loyola University Maryland, USA \\
$\ddagger$ Electrical and Computer Engineering Department, Tufts University, USA\\
 \texttt{\{vvtien,dat.duthanhcs\}@hcmut.edu.vn}, \texttt{tnho@loyola.edu}, \texttt{mai.vu@tufts.edu}
 }

\begin{document}

\maketitle


\begin{abstract}
Convolutional Neural Networks (CNNs) achieve high performance in image classification tasks but are challenging to deploy on resource-limited hardware due to their large model sizes. To address this issue, we leverage Mutual Information, a metric that provides valuable insights into how deep learning models retain and process information through measuring the shared information between input features or output labels and network layers. In this study, we propose a structured filter-pruning approach for CNNs that identifies and selectively retains the most informative features in each layer. Our approach successively evaluates each layer by ranking the importance of its feature maps based on Conditional Mutual Information (CMI) values, computed using a matrix-based Rényi $\alpha$-order entropy numerical method. We propose several formulations of CMI to capture correlation among features across different layers. We then develop various strategies to determine the cutoff point for CMI values to prune unimportant features. This approach allows parallel pruning in both forward and backward directions and significantly reduces model size while preserving accuracy. Tested on the VGG16 architecture with the CIFAR-10 dataset, the proposed method reduces the number of filters by more than a third, with only a $0.32\%$ drop in test accuracy.
\end{abstract}

\section{Introduction}

Convolution Neural Network (CNN) has achieved remarkable success in various tasks such as image classification, object detection, and segmentation \citep{zhang2019recent}, \citep{li2021survey}. Deeper architectures such as VGG16 \citep{simonyan_2014_very} and ResNet \citep{he_2016_deep} have shown superior performance in handling complex image classification tasks. However, the effectiveness of these networks is often reliant on very deep and wide architectures, resulting in a very large number of parameters that lead to longer training and inference time, and create challenges when deploying them on resource-constrained devices \citep{blalock2020state}, \citep{yang2017designing}.

CNNs often contain redundant weights and parameters, as certain weights learned in a network are correlated \citep{sainath2013low}.
To reduce network size and improve inference speed, network pruning techniques target different components such as weights, filters, and channels, using a range of criteria (see Related Work). A common approach is to measure the weight magnitudes to identify unimportant connections \citep{han2015deep}, \citep{molchanov2016pruning}, \citep{aghasi2020fast}. 

A less explored approach involves using mutual information between the network's output and latent features to detect redundant filters. \citet{yu2020understanding} assessed the information flow in CNNs by leveraging the Rényi $\alpha$-order entropy and conducted a preliminary analysis using Conditional Mutual Information (CMI) to identify key filters. {However, their study only uses CMI within a single layer, without considering the shared information among features across layers. Furthermore, the CMI-permutation method used to retain filters drastically underestimates the number of useful features. 
We confirmed in our experiments that the retained features in \citet{yu2020understanding} lead to a significant drop, of more than $80\%$, in model accuracy.}


In this paper, we build upon the concept of using CMI from \citet{yu2020understanding} to develop an effective method for pruning CNNs while preserving high accuracy. Our key contributions include advancing CMI computation across layers, defining optimal CMI cutoffs, and developing pruning strategies applicable to all CNN layers. Specifically, we introduce novel CMI formulations that capture shared information across multiple layers, improving the measure’s effectiveness in assessing feature importance. We also propose two methods for determining the CMI cutoff point to ensure optimal feature retention. Finally, we develop a robust algorithm for pruning CNN layers bidirectionally, starting from the most critical layer. Evaluations on the VGG16 architecture with the CIFAR-10 dataset demonstrate a $26.83\%$ reduction in parameters and a $36.15\%$ reduction in filters, with only a minimal $0.32\%$ drop in test accuracy, underscoring the effectiveness of our approach. 


\section{Related Work} 




Deep neural network pruning has seen major advancements in recent years, with various approaches on reducing model complexity while maintaining performance. These approaches can be categorized into pruning at initialization, dynamic pruning, unstructured pruning, and structured pruning.

\textit{Pruning at initialization} involves selecting weights or neurons likely to contribute little to the overall network performance and removing them { without using any gradient steps}. \citet{sadasivan2022ossum} designed OSSuM for pruning at initialization by applying a subspace minimization technique to determine which parameters can be pruned. 
\citet{tanaka2020pruning} proposed an approach to measure parameter importance called \emph{synaptic saliency} and ensured that {this metric} is preserved across layers. However, \cite{frankle2020pruning} critically examined popular pruning methods at initialization and argued that pruning during training remains more effective.


\textit{Dynamic pruning} approaches adjust the pruning process during training or inference. \citet{shneider2023impact} explored disentangled representations using the Beta-VAE framework, which enhances pruning by selectively eliminating irrelevant information in classification tasks. \citet{chen2023otov3} introduced OTOv3 that integrates pruning and erasing operations by leveraging automated search space generation and solving a novel sparse optimization. 

\textit{Unstructured pruning} removes individual weights rather than entire structures like filters, resulting in more flexibility but less hardware efficiency. \citet{molchanov2019importance} proposed a Taylor expansion-based pruning method that estimates the importance of weights by their impact on the loss function.
\citet{aghasi2020fast} introduced Net-Trim, which removes individual weights by formulating the pruning problem as a convex optimization to minimize 
\textcolor{black}{the sum of absolute entries of the weight matrices.} 
\citet{ding2019global} introduced Global Sparse Momentum SGD, a weight pruning technique that dynamically adjusts the gradient flow during training to achieve high compression ratios while maintaining model accuracy. \citet{lee2019signal} demonstrated the role of dynamical isometry in ensuring effective pruning across various architectures without prior training. 
\citet{han2015deep} combined weight pruning, quantization, and Huffman coding to achieve significant compression.

\textit{Structured Pruning} focuses on removing entire channels, filters, or layers, making it more compatible with modern hardware. \citet{he2023structured} provided a comprehensive survey in structured pruning of deep convolutional neural networks, emphasizing the distinction between structured and unstructured pruning and highlighting the hardware-friendly advantages of structured approaches. \citet{crowley2018closer} suggested that networks pruned and retrained from scratch achieve better accuracy and inference speed than pruned-and-tuned models.  \citet{you2019gate} developed the Gate Decorator method that employs a channel-wise scaling mechanism to selectively prune filters based on their estimated impact on the loss function, measured through a Taylor expansion. 
\citet{lin20221xn} grouped consecutive output kernels for pruning.
\citet{xu2019trained} integrated low-rank approximation into the training process, dynamically reducing the rank of weight matrices to compress the network. Considering Convolutional Neural Networks, various approaches have been introduced for filter pruning. \citet{guo2020dmcp} pruned filters using a differentiable Markov process to optimize performance under computational constraints; \citet{sehwag2020hydra} pruned filters based on an empirical risk minimization formulation; \citet{liu2019metapruning} utilized a meta-learning approach; \citet{molchanov2016pruning} interleaved greedy criteria-based pruning with fine-tuning by backpropagation, using a criterion based on Taylor expansion to minimize impact on the loss function. \citet{li2020eagleeye} developed EagleEye, a pruning method that leverages adaptive batch normalization to quickly and efficiently evaluate the potential of pruned sub-nets without extensive fine-tuning. 
\citet{he2017channel} proposed a channel pruning method based on LASSO regression and least squares reconstruction.
\citet{zhuang2018discrimination} incorporated additional discrimination-aware losses to maintain the discriminative power of intermediate layers.
\citet{he2019filter} proposed filter pruning via Geometric Median targeting redundant filters to reduce computational complexity.
\citet{yu2020understanding} proposed applying Conditional Mutual Information and Permutation-test to retain a set of important filters.

This paper shares a common objective with prior work in the \emph{structured pruning} domain, particularly focusing on filter pruning for Convolutional Neural Networks. While existing methods employ various pruning criteria, our study explores the application of mutual information (MI), specifically leveraging the matrix-based $\alpha$-order Rényi entropy computation to produce MI values which are used to guide the pruning process. This paper contributes to the area of applying MI in machine learning, emphasizing the use of MI to identify and retain the most informative filters across layers.

\section{Computing the CMI Values of Candidate Feature Sets} \label{sec:4}


In this section, we analyze the use of Conditional Mutual Information (CMI) as a metric to measure feature importance, and discuss several approaches to ordering the features in each CNN layer and computing their CMI values. We propose new CMI computation that leverages shared information across layers and further exploit Markovity between layers to make the computation efficient.

\subsection{Selected Features Set and Non-selected Features Set} We first define the notation used for the rest of the paper. Let $\displaystyle \mX$ and $\displaystyle \mY$ be the input and output data of the CNN. We consider a pretrained CNN model that has $N$ CNN layers, $\{L_i\}_{i=1,...,N}$. Each layer $L_i$ contains multiple feature maps obtained by feed-forwarding the training data to this layer using the layer filters. At each layer $L_k$, the feature map selection process involves separating the set of feature maps $F_k$ at layer $L_k$ into two distinct sets: the selected set $F_k^s$ and the non-selected set $F_k^n$, that is, $F_k = \{F_k^s, F_k^n\}$.

\textbf{Selected feature set} $F_k^s$ is a subset of the feature map set $F_k$ at layer $L_k$ and consists of feature maps selected according to a selection criterion as discussed later in Section \ref{sec:cutoff}. 
The selection criteria are designed to retain a high test accuracy on the retrained CNN model after pruning.

\textbf{Non-selected feature set} $F_k^n$ is the rest of the feature maps at layer $L_k$, i.e. $F_k^n = F_k\setminus F_k^s$, which consists of feature maps that do not significantly contribute to the model's performance, and hence can be pruned to simplify the model complexity without compromising accuracy.

\textbf{Selection metric:} We are interested in the information that the feature maps in each layer convey about the CNN output, which can be measured by the mutual information (MI) between the feature map set $F_k$ and the output $Y$. Note the following MI relationship:
\begin{equation}
    I(Y; F_k) = I(Y; F_k^s, F_k^n) = I(Y;F_k^s) + I(Y;F_k^n | F_k^s)
    \label{eq:mi_relation}
\end{equation}
We observe that the selected feature set $F_k^s$ will convey most information about the output $Y$ if the second term of the summation in Eq. (\ref{eq:mi_relation}) is sufficiently small. This second term measures the conditional mutual information (CMI) between the non-selected feature set and the output, conditioned on the selected feature set. That is to say, \emph{given the selected feature set $F_k^s$, if the non-selected feature set $F_k^n$ does not bring much more information about the CNN output, then it can be effectively pruned without affecting CNN accuracy performance}. As such, in our algorithms, we will compute the CMI values of various candidate feature sets for pruning to determine the best set to prune. 


\subsection{Ordering Features With Per-layer Conditional Mutual Information}
\label{sec:per_layer_cmi}



We now discuss how to use conditional mutual information (CMI) 
to rank the feature maps in each CNN layer. 
The ordered list based on CMI values will later be used for pruning. 
Here we review the method for ordering features and computing CMI values within one layer as in \citep{yu2020understanding}; in the next section, we propose new methods for ordering features and computing CMIs across layers.

\textbf{Ordering features per layer:   } 
Consider layer $L_k$ with the set of feature maps $F_k$ in a pre-trained CNN. To order the feature maps in $F_k$, we compute the MI 
between each unordered feature map and the output $Y$, then incrementally select the one that maximizes the MI. Specifically, starting from an empty list of ordered features $F_{k}^{o} = [\emptyset]$ and a full list of non-ordered features  $F_{k}^{u} = F_k$, we successively pick the next best feature map $f^\star$ from $F_k^{u}$ that maximizes \citep{yu2020understanding}
\begin{equation} \label{eq:ordering_per_layer}
    f^\star = \underset{f \in F_{k}^{u}} {\text{argmax}} \; 
    I(Y;  F_{k}^{o} \cup \{ f \} ) \hspace{0.02in}.
\end{equation}
Once the next best feature map $f^\star$ is identified, it is moved from the unordered feature list $F_k^{u}$ to the ordered feature list $F_k^{o}$ as follows.
\begin{equation} \label{eq:set_update}
    F_{k}^{o} = F_{k}^{o} \cup \{ f^\star \}; \hspace{0.05in}
    F_{k}^{u} = F_{k}^{u} \setminus \{ f^\star \} \hspace{0.02in}.
\end{equation}
This process is repeated iteratively for $|F_k|$ times to order all the feature maps of layer $L_k$.


\textbf{Computing the per-layer CMI values:   }
Each time the two lists are updated with a newly ordered feature map as in Eq. (\ref{eq:set_update}), they create new candidates for feature selection, where $F_k^o$ is a candidate for the selected feature set, and $F_k^u$ for the non-selected feature set.
To evaluate the "goodness" of these candidate sets, we compute the CMI at each ordering iteration $i$ as follows \citep{yu2020understanding}.
\begin{equation} \label{eq:cmi_wi_1layer}
    c_i = I(Y; F_{k,i}^{u} | F_{k,i}^{o}) \;, \quad i = 1 \ldots |F_k|
\end{equation}
where index $i$ refers to the $i$-th iteration of performing ordering steps  (\ref{eq:ordering_per_layer}) and (\ref{eq:set_update}) in layer $L_k$.

As $i$ increases, the ordered feature list $F_{k,i}^{o}$ grows and the non-order feature list $F_{k,i}^{u}$ shrinks, hence the value of $c_i$ is automatically decreasing with $i$. At the end of this process, each CNN layer will have an associated list of decreasing CMI values $C_k = \{c_1, c_2, \ldots, c_{n_k}\}$, where $n_k=|F_k|$.  


\subsection{Ordering Features With Cross-layer Conditional Mutual Information} 
The above per-layer CMI computation ignores shared information among features across different layers. To utilize this cross-layer relation, we consider cross-layer CMI computations that incorporate information from multiple CNN layers into the pruning process of each layer. We propose two methods for ordering the features of each layer and computing the cross-layer CMI values.

\subsubsection{Full CMI conditioned on all previously considered layers} 
\label{sec:full_cmi}
We follow a similar process as above 
but replace the maximization criterion in (\ref{eq:ordering_per_layer}) with (\ref{eq:mi_cross_layer_full}), and the CMI computation in (\ref{eq:cmi_wi_1layer}) with (\ref{eq:cmi_cross_layer_full}) below. Specifically, let $F_{1}^{s},F_{2}^{s}, \ldots,F_{k-1}^{s}$ be the lists of selected feature maps of previously explored CNN layers $L_1, \ldots, L_{k-1}$. 
At layer $L_k$, the next feature $f^\star$ to be added to the ordered list $F_{k}^{o}$ will be chosen as 
\begin{equation} \label{eq:mi_cross_layer_full}
    f^\star = \underset{f \in F_{k}^{u}} {\text{argmax}} \; I(Y; F_{1}^{s},\ldots,F_{k-1}^{s}, F_{k}^{o} \cup \{ f \} )
\end{equation}
After updating the ordered list with the new feature map $f^\star$ as in Eq. (\ref{eq:set_update}), we calculate the CMI value of the new unordered set as 
\begin{equation} \label{eq:cmi_cross_layer_full}
   c = I(Y; F_{k}^{u} | F_{1}^{s},\ldots,F_{k-1}^{s}, F_{k}^{o})
\end{equation}
Steps (\ref{eq:mi_cross_layer_full}), (\ref{eq:set_update}), and (\ref{eq:cmi_cross_layer_full}) are repeated $|F_k|$ times for each layer $L_k$. At the end of this process, each layer again has a list $C_k$ of decreasing CMI values. 


\subsubsection{Compact CMI conditioned on only the last layer} 
\label{sec:compact_cmi}
In feedforward Deep Neural Networks inference, input signals are propagated forward from the input layer to the output layer, passing through multiple hidden layers. In each propagation, the computation flows in a single direction, with the latent features at each layer depending only on the signals from the previously considered layer and weights of the current layer, hence forming a Markov chain \citep{yu_2019_Understanding}. The Markov property implies that the CMI values computed at a certain layer depend solely on the immediately preceding or succeeding layer \citep{cover1999elements}. We stress that this Markov property applies in both directions for CMI computation, whether the given sets that are being conditioned on come from the preceding layers or succeeding layers. (This is because of the property that if $X \rightarrow L \rightarrow Y$ forms a Markov chain, then $Y \rightarrow L \rightarrow X$ also forms a Markov chain.) We will later exploit this property to design pruning algorithms that work in both directions. For the easy of exposition, however, we will only show the forward CMI computation here, but noting that it can be applied in the backward direction as well.

Leveraging the Markovity among layers, we propose a more compact method for computing cross-layer CMI values at each layer $L_k$. This method replaces
steps (\ref{eq:mi_cross_layer_full}) and (\ref{eq:cmi_cross_layer_full}) with 
(\ref{eq:mi_cross_layer_compact})
and (\ref{eq:cmi_cross_layer_compact}) respectively as below. The feature ordering maximization criterion becomes 
\begin{equation} \label{eq:mi_cross_layer_compact}
    f^\star = \underset{f \in F_{k}^{u}} {\text{argmax}} \; I(Y; F_{k-1}^{s}, F_{k}^{o} \cup \{ f \} )
\end{equation}
and the compact CMI computation used to create the CMI list is 
\begin{equation} \label{eq:cmi_cross_layer_compact}
    c = I(Y; F_{k}^{u} | F_{k-1}^{s}, F_{k}^{o})
\end{equation}
Steps (\ref{eq:mi_cross_layer_compact}), (\ref{eq:set_update}), and (\ref{eq:cmi_cross_layer_compact}) are repeated $|F_k|$ times for each layer $L_k$ to produce the CMI list $C_k$.

\subsubsection{Full CMI versus Compact CMI and Examples}
While the compact CMI in (\ref{eq:cmi_cross_layer_compact}) and the full CMI in (\ref{eq:cmi_cross_layer_full}) are theoretically equivalent because of Markovity among CNN layers, their numerical values may vary in practice due to the estimation methods used for calculating mutual information and the numerical precision of the machine. Specifically, we use the matrix-based numerical method for computing Rényi entropy in (\ref{eq:Renyi_matrix_based}) (see Appendix) from layer data without having the true distributions, thus the computed values for compact CMI and full CMI diverge when conditioned on more layers. Therefore, we conduct an ablation study to compare both approaches in the experimental evaluation presented in Section \ref{sec:experiment}.

\begin{wrapfigure}{rb}{0.55\textwidth}
\begin{center}
\includegraphics[width=0.55\textwidth]{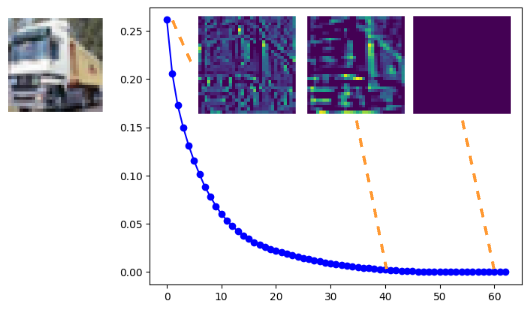}
\caption{Example of ordered feature maps using cross-layer compact CMI computation in Alg. \ref{alg:cmi_computation}. The top left figure is the input image with label \textit{truck}. The vertical axis presents the computed CMI value and the horizontal axis shows the index of the newly added ordered feature map.}
\label{fig:example_ordered_features}
\end{center}
\end{wrapfigure}
Algorithm \ref{alg:cmi_computation} 
provides the implementation details of feature ordering and CMI computation for all three methods: per-layer CMI, cross-layer full CMI, and cross-layer compact CMI. The algorithm returns the fully ordered feature set $F_k^o$ of layer $L_k$ and the set of decreasing CMI values $C_k$.



Figure \ref{fig:example_ordered_features} provides an example illustrating the ordered feature maps in a CNN layer based on cross-layer compact CMI values. This particular CNN layer has 64 feature maps, whose indices are shown on the horizontal axis in the order of decreasing CMI values as shown on the vertical axis. At index points 1, 40, and 60, we display the corresponding newly added feature map to the ordered feature set. The first feature map shows a relatively clear pattern related to the input image of a \textit{truck}, while the middle one becomes more blurry, and the last feature map does not at all resemble the truck. In the next section, we present two different approaches, Scree test and X-Mean clustering, for selecting a cutoff point to prune the feature maps based on CMI values. Using these approaches, the added feature maps at points 1 and 40 are retained, whereas the feature map at point 60 is consistently pruned. This means the set of last five feature maps from 60 to 64 contains little information about the CNN output and can be pruned without affecting accuracy performance.

\begin{algorithm}[t]
\caption{Feature ordering with CMI Computation}
\label{alg:cmi_computation}
\begin{algorithmic}[1]
\STATE \textbf{Input:} Selected features set $F_{1}^{s}, F_{1}^{s},\ldots,F_{k-1}^{s}$ of layer $L_{1},L_{2},\ldots,L_{k-1}$, full feature set of current layer $F_k$, output $Y$
\STATE \textbf{Initialize:} $F_{k}^{o} = [\;\emptyset \;]$, $F_{k}^{u} = F_k$, $C_k = [\;\emptyset \;]$
\WHILE{$|F_{k}^{u}| \geq 1$}
    \STATE Find 
    $\; f^\star \;$ according to Eq. (\ref{eq:ordering_per_layer}) or Eq. (\ref{eq:mi_cross_layer_full}) or Eq. (\ref{eq:mi_cross_layer_compact})
    \STATE Update:
    $\; F_{k}^{u} = F_{k}^{u} \setminus \{ f^\star \}; \quad  F_{k}^{o} = F_{k}^{o} \cup \{ f^\star \}$
    \STATE Compute CMI value 
    $\; c  \;$ according to Eq. (\ref{eq:cmi_wi_1layer}) or Eq. (\ref{eq:cmi_cross_layer_full}) or Eq. (\ref{eq:cmi_cross_layer_compact}), respectively as in Step 4

    \STATE Append $\; C_k = \{C_k, c\}$
\ENDWHILE
\STATE \textbf{return} $F_{k}^{o}$, $C_k$
\end{algorithmic}
\end{algorithm}

\section{Determining a Cutoff Point for CMI Values in Each Layer}
\label{sec:cutoff}

After ordering the features of each CNN layer and computing the CMI values of candidate sets of features as in Section \ref{sec:4}, the features are arranged in descending order of CMI values. The next step is to determine a cutoff point within the ordered list of CMI values such that the set of features with CMI value at the cutoff point is selected and retained, and the set of features with lower CMI, which contributes little to the CNN output, is pruned. In this section, we propose two methods to identify such a cutoff point based on the Scree test and X-Mean clustering.


\subsection{Identifying Cutoff Point using Scree Test} 
The Scree test \citep{cattell_1966_meaning} is first proposed in Principal component analysis (PCA) to determine the number of components to be retained using their eigenvalues plotting against their component numbers in descending order. The point where the plot shifts from a steep slope to a more gradual one indicates the meaningful component, distinct from random error \citep{da_2005_scree}. Furthermore, \citet{niesing_1997_simultaneous} introduced the Quotient of Differences in Additional values (QDA) method, which identifies the $q^{th}$ component that maximizes the slope $s(q) = (\lambda_q - \lambda_{q+1})(\lambda_{q+1} - \lambda_{q+2})^{-1}$
where $\lambda_q$ is the eigenvalue for the $q^{th}$ component in PCA.

Here we apply the QDA method \citep{niesing_1997_simultaneous} to the list of decreasing CMI values obtained as in Section \ref{sec:4}. 
To explore more than one candidate cutoff point, we propose to find $K$ CMI values that correspond to the top $K$ largest slopes as
\begin{equation}
\{i_1, i_2, \ldots, i_K \} = 
\underset{{i=1 \ldots |F_k|-2}} {\text{top K}} \; \frac{c_{i} - c_{i+1}}{c_{i+1}-c_{i+2}} ,
\label{eq:cutoff_topk_scree}
\end{equation}
Each of the $K$ candidate cutoff points from the list obtained above will be examined by carrying out trial pruning of current layer $L_k$ (pruning off the set of features beyond each point) and testing the resulting pruned model for accuracy. {(This pruned model is the one obtained right at this pruning step in the current layer and is not the final pruned model.)} 
The optimal cutoff point will then be chosen based on the resulting pruned model's accuracy while maximizing the pruning percentage. Specifically, denote $a^{f}$, $a^{p}$ as the accuracy of the full and pruned models, respectively, and let $\delta$ be the targeted maximum reduction in accuracy such that $a^{f} - a^{p} \leq \delta$. Then the optimal cutoff point is the one from (\ref{eq:cutoff_topk_scree}) which results in the largest pruned percentage while satisfying the accuracy requirement. If no candidate point meets this accuracy threshold, the index with the highest accuracy is chosen. Since this process involves trial pruning and testing for accuracy of the pruned model, typically only a small value of $K$ is used, around 2 or 3 cutoff point candidates. In the special case of $K=1$, only the cutoff point with maximum slope is chosen and no trial pruning is necessary. Algorithm \ref{alg:prune_with_scree_test} outlines the procedure for selecting the optimal cutoff point using the Scree test. 

\begin{algorithm}[t]
\caption{Determining CMI Cutoff Point in a Layer Using Scree Test}
\label{alg:prune_with_scree_test}
\begin{algorithmic}[1]
\STATE \textbf{Input:} List of ordered features $F^{o}_k$ from layer $L_k$, list of CMI values $C_k$, pre-trained CNN model $M$, training dataset $D$, target accuracy threshold $a^p$, number of top candidates $K$

\STATE \textbf{Initialize:} List of cut-off index and model accuracy $A = [\hspace{0.05in}]$
\FOR{$i = 1$ to $(|C_k|-2)$}
    \STATE $s(i) = \frac{c_i - c_{i+1}}{c_{i+1} - c_{i+2}}$ \hspace{0.1in }// \textit{Compute QDA score for each CMI point}
\ENDFOR

\STATE \textbf{Find} top $k$ largest $s(i)$ values and their corresponding indices $\{i_1, i_2, \ldots, i_K\}$

\FOR{$j = 1$ to $K$}
    \STATE 
    Prune all features in $F_k^o$ with indices after $i_j$ to obtain an intermediate pruned model $M_j$
    \STATE Evaluate $M_j$ on $D$ to obtain the accuracy $a_j$
    \STATE Append $(i_j, a_j)$ to $A$
\ENDFOR
\STATE Choose the smallest index $i^\star$ with $a_{i^\star} \geq a^p$ or else $i^\star = \max\{i_1, \ldots, i_K\}$
\STATE Select all features up to index $i^\star$, and prune all features after $i^\star$ in $F_k^o$, to obtain $F^{s}_k$
\STATE \textbf{return} $(i^\star, F^{s}_k)$
\end{algorithmic}
\end{algorithm}

\subsection{Identifying Cutoff Point using X-Means Clustering} 

Here we propose an alternative method to select the optimal CMI cutoff point based on clustering using the X-Means algorithm \citep{pelleg_2000_xmeans}, an extension of $k$-means, to cluster the CMI values into different groups. X-Means automatically determine the optimal number of clusters based on the Bayesian Information Criterion
\textcolor{black}{$\text{BIC}(M) = \mathcal{L}(D) - \frac{p}{2}\log(R)$ where $\mathcal{L}(D)$ is the log-likelihood of dataset $D$ with $R$ samples according to model $M$ with $p$ parameters.
}



\begin{wrapfigure}{r}{0.4\textwidth}
\includegraphics[width=0.38\textwidth]{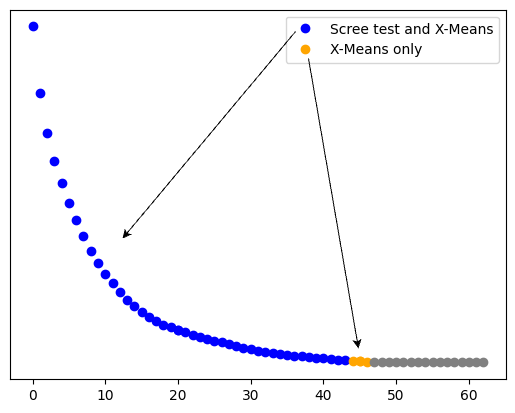}
\caption{Example of cutoff points by Scree test and X-Means.}
\label{fig:examples-scree-xmeans}
\end{wrapfigure}
X-Means starts with an initial cluster number, and increases this number until the BIC score stops improving. Once clusters are formed in the current layer, {we order the clusters based on the CMI value of the cluster center point in decreasing order.} Starting with the first cluster, we retain all its feature maps and perform trial pruning of the remaining feature maps from all other clusters. The pruned model's accuracy is then evaluated. As the process continues, new features from the next cluster are added to the selected feature set, until the test accuracy meets or exceeds the targeted accuracy threshold. Algorithm \ref{alg:prune_with_xmeans} provides the outline of this X-Means procedure. 

\begin{algorithm}[t]
\caption{Determining CMI Cutoff Point in a Layer using X-Means Clustering}
\label{alg:prune_with_xmeans}
\begin{algorithmic}[1]
\STATE \textbf{Input:} List of ordered feature maps $F^o_k$ of layer $L_k$, list C of CMI values, pre-trained model $M$, training dataset $D$, accuracy threshold $a^{p}$

\STATE Apply X-means on $C$ to obtain $K$ clusters $e_1, \ldots, e_K$, ordered in the decreasing CMI value of the cluster center
\STATE \textbf{Initialize}: A = [\hspace{0.05in}], $F^s_k = F^o_k$
\FOR{$j = 1$ to $K$}
\STATE Append features in $e_j$ to A
	\STATE Prune features in $e_{j+1}$ \text{to} $e_K$ from model $M$ to obtain an intermediate pruned model $M_j$
	\STATE Evaluate $M_j$ on $D$ to obtain accuracy $a_j$
\STATE {\bf If} $a_j >= a^{p}$ {\bf then} $F^s_k$ $\leftarrow$ A and {\bf break}
\ENDFOR

\STATE \textbf{return} $F^s_k$
\end{algorithmic}
\end{algorithm}

Figure \ref{fig:examples-scree-xmeans} illustrates the cutoff points selected by using the Scree test and X-Means clustering methods. We see that the majority of feature maps selected by the Scree test and X-Means clustering are similar\textcolor{black}{, represented by the blue points}. \textcolor{black}{The orange points indicate feature maps retained only by X-Means, and the gray points represent feature maps pruned by both methods.} The difference between the two methods boils down to only the last few feature maps. In this example, the Scree test retains 43 while X-Means retains 46 out of the total 64 feature maps. 


\section{Algorithms for Pruning All Layers of a CNN based on CMI}

We now combine methods from the previous two sections in an overall process to systematically traverse and prune every layer of a CNN. We propose two algorithms that differ in their starting layer and pruning direction. One algorithm begins at the first convolutional layer and prunes forward through the network. The other algorithm starts at the layer with the highest per-layer pruning percentage and simultaneously prunes both forward and backward from there.

\begin{figure}[t]
\begin{center}
\includegraphics[width=\textwidth]{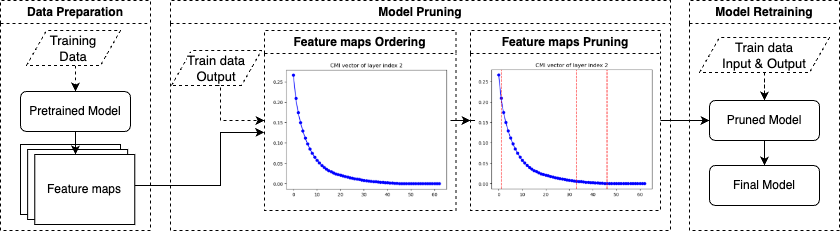}
\end{center}
\caption{Overview of the CMI-based pruning process. The blue curve shows a list of decreasing CMI values as new feature maps are sequentially added to the order set of each layer. The red vertical lines indicate candidate cutoff points for the CMI list. The important feature maps to be selected and retained are those to the left of the red lines.} 
\label{fig:cnn_pruning_overview}
\end{figure}

The pruning process consists of three phases as illustrated in Figure \ref{fig:cnn_pruning_overview}. The first phase is \emph{Data Preparation} which generates the feature maps of each layer. We start with a pre-trained CNN model that feeds forward the data using mini-batch processing through each CNN layer $L_k$ to produce a set of feature maps $F_k$. 
The second stage is the main \emph{Pruning Algorithm} in which every convolutional layer of the CNN is processed and pruned in a certain order. The last stage is \emph{Retraining} of the pruned model to fine-tune the model parameters to improve accuracy performance.

\subsection{Forward Model Pruning}
In \textit{{Forward Pruning}}, the algorithm starts with the first convolutional layer and prunes all convolutional layers sequentially from first to last. At each layer, the algorithm applies the chosen feature ordering and CMI computation method (Section \ref{sec:4}) to produce the decreasing CMI value list, then applies the chosen cutoff point identification method (Section \ref{sec:cutoff}). In cross-layer CMI computation, the CMI values of each layer are computed by conditioning on the selected feature sets of previous layers. Algorithm \ref{alg:forward_cmi} describes this forward pruning procedure. 

\begin{algorithm}[t]
\caption{Forward Pruning Procedure} 
\label{alg:forward_cmi}
\begin{algorithmic}[1]
\STATEx \textbf{Input:} Set of feature maps $\{F_1, F_2, \ldots, F_N\}$, output $Y$,
pre-trained CNN model $M$, model accuracy $a^f$, accuracy threshold $a^p$, training data $D$ 

\FOR{$k = 1$ to $N$}
    \STATE $C_k, F_{k}^{o}$ $\leftarrow$ \text{Rank features in $L_k$ using} \textit{cross-layer CMI} (Alg. \ref{alg:cmi_computation}) \text{with inputs} $F^{s}_{k-1}, F_k, Y$
    
    \STATE $F_k^s$ $\leftarrow$ \text{Find cutoff point within} \textit{CMI list} (Alg. \ref{alg:prune_with_scree_test} or \ref{alg:prune_with_xmeans}) \text{with inputs} $C_k, F_k^{o}, M, D, a^p$
    

\ENDFOR
\STATE \textbf{return} selected feature set for each layer $F_1^s,\ldots,F_N^s$
\end{algorithmic}
\end{algorithm}

\subsection{Bi-directional Model Pruning}

We design {\textit{Bi-directional Pruning}} to improve the previous pruning approach by first determining the most effective layer to begin the pruning process. We propose to start with the layer that has the highest per-layer pruning percentage while maintaining an acceptable post-pruning accuracy. First, we perform trial-pruning of each convolutional layer of the CNN individually, using \emph{per-layer CMI computation} and either the Scree test or X-Means method. This initial stage lets us identify the layer with the highest pruning percentage as the starting layer for the full CNN pruning process. Next, we start from the identified best layer and proceed by using \emph{cross-layer CMI computation} to prune the original CNN in both directions, forward and backward. For \emph{compact CMI} computation, at each new layer, the compact CMI values are conditioned on the immediately previous layer that was pruned, which can either be the preceding layer (in forward pruning) or the succeeding layer (in backward pruning). For \emph{full CMI} computation, we condition the CMI on all previously pruned layers from the starting layer in the corresponding direction. We note that in Bi-directional pruning, per-layer CMI computation in Eq. (\ref{eq:cmi_wi_1layer}) is only used at the initial stage to determine the starting layer; after that, the pruning process uses cross-layer CMI computation in Eq. (\ref{eq:cmi_cross_layer_compact}) or Eq. (\ref{eq:cmi_cross_layer_full}). Algorithm \ref{alg:multi_level_cmi} outlines the detailed procedure of Bi-directional Pruning. 

\begin{algorithm}[t]
\caption{Bi-directional Pruning Procedure} 
\label{alg:multi_level_cmi}
\begin{algorithmic}[1]
\STATEx \textbf{Input:} Set of feature maps $\{F_1, F_2, \ldots, F_N\}$, output $Y$, pre-trained model $M$, accuracy of pre-trained model $a^f$, accuracy threshold $a^p$, training data $D$


\FOR{$k = 1$ to $N$} 
	\STATE $C_k, F_{k}^{o}$ $\leftarrow$ \text{Compute} \textit{per-layer CMI} (Alg. \ref{alg:cmi_computation})  
	\STATE $F_k^s$, $a_k$ $\leftarrow$ \textit{Prune $(C_k, F_k^o)$ using Scree test or X-Means} (Alg. \ref{alg:prune_with_scree_test} or \ref{alg:prune_with_xmeans})
	\STATE $r_{k} \leftarrow 1 - |F_k^s| / |F_k|$ \hspace{0.1 in}// \textit{pruning ratio $r_k$}
\ENDFOR
\STATE \text{Determine} layer $k^\star$ with the highest pruning percentage $r_{k^\star}$ and $a_{k^\star} >= a^p$ 
\STATEx \textit{// Forward CMI Computation}
\FOR{$k = k^\star$ + 1 \textbf{to} $N$}
    \STATE $C_k, F_{k}^{o}$ $\leftarrow$ \text{Compute} \textit{cross-layer CMI} values (Alg. \ref{alg:cmi_computation}) for layer $L_k$ 
    
    \STATE $F_k^s$ $\leftarrow$ \textit{Prune $(C_k, F_k^o)$ using Scree test or X-Means} (Alg. \ref{alg:prune_with_scree_test} or \ref{alg:prune_with_xmeans})
\ENDFOR
\STATEx \textit{// Backward CMI Computation} 
\FOR{$k = k^\star$ - 1 \textbf{down to} $1$}
    \STATE $C_k, F_{k}^{o}$ $\leftarrow$ \text{Compute} \textit{cross-layer CMI} values (Alg. \ref{alg:cmi_computation}) for layer $L_k$ 
    
    \STATE $F_k^s$ $\leftarrow$ \textit{Prune $(C_k, F_k^o)$ using Scree test or X-Means} (Alg. \ref{alg:prune_with_scree_test} or \ref{alg:prune_with_xmeans})

\ENDFOR
\STATE \textbf{return} Set of selected features for each layer $F_1^s,\ldots,F_N^s$
\end{algorithmic}
\end{algorithm}

\section{Experimental Results}\label{sec:experiment}
This section presents our experimental evaluation of the CNN pruning algorithms. Due to space, we present the main results here and delegate detailed results and ablation studies to the Appendix.

\subsection{Experiment Setup}

We evaluate our proposed pruning algorithms on VGGNet \citep{simonyan_2014_very}, specifically a VGG16 model which consists of 13 convolutional layers \citep{huy_phan_2021_4431043}, pre-trained on 
the CIFAR-10 dataset \citep{krizhevsky_2009_learning}. 
We use the training data to evaluate the accuracy of the intermediate pruned models, and the test data to evaluate the accuracy of the final pruned model.
When preparing the data, we use a batch of 256 training samples to feed forward through the VGG16 model and generate the feature maps at each layer for use in our algorithms.

We performed several experiments to prune the original CNN model using different combinations of CMI computation and cutoff point methods as in Algorithms \ref{alg:forward_cmi} and \ref{alg:multi_level_cmi}. When using the Scree-test with multiple candidates, we set $K=3$. 
The original accuracy on training data is $99.95\%$ \citep{huy_phan_2021_4431043}, and to check the accuracy of the intermediately pruned models, 
we set the target accuracy as $a^{p} = 98.95\%$. In all experiments in this section, \emph{we prune the CNN model by completely removing the weights corresponding to the pruned features in each layer (Actual pruning -- see Appendix)}. The final convolutional layer is not pruned to maintain all connections to the first fully connected layer. The pruning efficiency is determined by the percentage of pruned filters over all filters.

After the CNN model is fully pruned, we \emph{re-train each pruned model} to fine-tune the weights for better test accuracy. For the retraining process, we apply the VGG16 training parameters for CIFAR-10 as in \citep{huy_phan_2021_4431043} and train each pruned model with 100 epochs.

\subsection{Analysis of Feature Maps Ordering and CMI Computation Methods}

Table \ref{tab:fm-ordering-comparison} shows a comparative analysis of the various feature maps ordering and CMI computation approaches as discussed in Section \ref{sec:4} (Algorithms \ref{alg:cmi_computation}). The cutoff point selection method in this set of experiments is the Scree-test. The results are displayed in terms of the number of retained parameters, pruned percentage of filters, and test accuracies before and after retraining. 

The Bi-directional pruning algorithm with cross-layer compact CMI computation (Algorithm \ref{alg:cmi_computation}) yields the smallest pruned model size (24.618 M parameters retained), representing $26.84\%$ parameter reduction from the original model. The same algorithm also results in the highest \textit{pruned percentage} of $36.15\%$ filters removed. Although this most aggressive pruning approach leads to a slightly lower accuracy \emph{before retraining} compared to other approaches, it actually achieved the best test accuracy \emph{after retraining}. After retraining, all considered methods converged to a similar accuracy. The original model's test accuracy was $94\%$, and after retraining for 100 epochs, this most aggressively pruned model achieves a test accuracy of $93.68\%$, which is the best among all experimented methods. \emph{This result confirms the validity of our approach of using cross-layer compact CMI computation and pruning in both directions.} 


\begin{table}
\caption{CNN Pruning using Scree-test Cutoff Point with various CMI Computation Methods}
\label{tab:fm-ordering-comparison}
\begin{center}
\begin{tabular}{|p{5.5cm}|p{1.65cm}|p{1.65cm}|p{1.65cm}|p{1.65cm}|}
\hline
\bf \vspace{0.15cm} CNN Pruning Algorithms & \bf \vspace{0.01cm} Parameters Retained & \bf Filters Pruned Percentage & {\bf Accuracy before Retraining} & {\bf Accuracy after  \hspace{0.5cm} Retraining}
\\
\hline 
\it{No pruning (original model)} & \textit{\textbf{33.647 M}} & \textit{ \textbf{0 \%}} & \textit{ \textbf{94.00\%}} & -- \\
{Forward pruning \& full CMI} & 33.196 M & 2.18\% & 93.02\% & 93.67\% \\
{Forward pruning \& compact CMI} & 25.7 M  & 26.70\% & 90.17\% & 93.33\% \\
Bi-directional pruning \& full CMI & 25.643 M & 30.12\% & 88.59\% & 93.25\% \\
Bi-directional pruning \& compact CMI & \bf 24.618 M & \bf 36.15\% & 90.95\% & \bf 93.68\% \\
\hline 
\end{tabular}
\end{center}
\end{table}

\subsection{Analysis of CMI Cutoff Point Approaches}

In this set of experiments, we compare the two proposed CMI cutoff point approaches, Scree-test and X-means, with the Permutation-test in  \citep{Yu2021}. For Permutation-test, we use a permutation number of 100 and a significance level of $0.05$ as used in \citep{Yu2021}. The CNN pruning algorithm is Bi-directional Pruning with Cross-layer Compact CMI computation (Alg. \ref{alg:multi_level_cmi}). Table \ref{tab:fm-ordering-comparison} shows the effectiveness of different cutoff point approaches when applied to the VGG16 model. 

The Permutation-test \citep{Yu2021} shows the smallest pruned model size but at a drastically reduced test accuracy to only $10.02\%$ even after retraining. This shows that the Permutation test was not able to differentiate unimportant features from the important ones and hence pruned aggressively and indiscriminately. In contrast, the proposed Scree-test and X-means both achieve more than a third of the features pruned while still retaining most of the accuracy of the original model.
The results show that Scree-test is slightly more robust than X-means by achieving both a higher pruned percentage and a better retrained-accuracy. This could be because Scree-test is more effective at preserving the most important feature maps compared to X-means.



\begin{table}
\caption{Bi-directional Pruning with Compact CMI using Various Cutoff Point Approaches}
\label{tab:fm-pruning-comparison}
\begin{center}
\begin{tabular}{|p{4.7cm}|p{1.65cm}|p{1.65cm}|p{1.65cm}|p{1.65cm}|}
\hline
\bf \vspace{0.03cm} Cutoff Point Approaches & \bf \vspace{0.01cm} Parameters Retained & \bf Filters Pruned Percentage & {\bf Accuracy before Retraining} & {\bf Accuracy after \hspace{0.5cm} Retraining}
\\
\hline 
\it{No pruning (original model)} & 33.647 M & 0 \% & 94.00\% & - \\
{Permutation-test \citep{Yu2021}} & 19.379 M & 81.79\% & 9.99\% & 10.02\% \\
{Scree-test} & \bf 24.618 M & \bf 36.15\% & \bf 90.95\% & \bf 93.68\% \\
{X-means} & 25.01 M & 34.67\% & 83.56\% & 92.99\% \\
\hline 
\end{tabular}
\end{center}
\end{table}

\section{Conclusion}

In this study, we introduced novel structured pruning algorithms for Convolutional Neural Networks (CNNs) by using Conditional Mutual Information (CMI) to rank and prune feature maps. By applying matrix-based Rényi $\alpha$-order entropy computation, we proposed several CMI-based methods for identifying and retaining the most informative features while removing redundant ones. Two different strategies, Scree test and X-means clusterng, were explored to determine the optimal cutoff points for pruning. We also examine both forward and backward prunings which were found to be effective. Our experiments demonstrated that the proposed approach significantly reduces the number of parameters by more than a third with negligible loss in accuracy, achieving efficient model compression. This method provides a promising framework for deploying CNN models on resource-constrained hardware without compromising performance. Future work may explore extending this approach to other network architectures and tasks beyond image classification.


\newpage
\bibliography{iclr2025_conference}

\begin{thebibliography}{50}
\providecommand{\natexlab}[1]{#1}
\providecommand{\url}[1]{\texttt{#1}}
\expandafter\ifx\csname urlstyle\endcsname\relax
  \providecommand{\doi}[1]{doi: #1}\else
  \providecommand{\doi}{doi: \begingroup \urlstyle{rm}\Url}\fi

\bibitem[Aghasi et~al.(2020)Aghasi, Abdi, and Romberg]{aghasi2020fast}
Alireza Aghasi, Afshin Abdi, and Justin Romberg.
\newblock Fast convex pruning of deep neural networks.
\newblock \emph{SIAM Journal on Mathematics of Data Science}, 2\penalty0 (1):\penalty0 158--188, 2020.

\bibitem[Blalock et~al.(2020)Blalock, Gonzalez~Ortiz, Frankle, and Guttag]{blalock2020state}
Davis Blalock, Jose~Javier Gonzalez~Ortiz, Jonathan Frankle, and John Guttag.
\newblock What is the state of neural network pruning?
\newblock \emph{Proceedings of machine learning and systems}, 2:\penalty0 129--146, 2020.

\bibitem[Cattell(1966)]{cattell_1966_meaning}
Raymond~B Cattell.
\newblock The meaning and strategic use of factor analysis.
\newblock In \emph{Handbook of multivariate experimental psychology}, pp.\  131--203. Springer, 1966.

\bibitem[Chen et~al.(2023)Chen, Ding, Zhu, Chen, Wu, Zharkov, and Liang]{chen2023otov3}
Tianyi Chen, Tianyu Ding, Zhihui Zhu, Zeyu Chen, HsiangTao Wu, Ilya Zharkov, and Luming Liang.
\newblock Otov3: Automatic architecture-agnostic neural network training and compression from structured pruning to erasing operators.
\newblock \emph{arXiv preprint arXiv:2312.09411}, 2023.

\bibitem[Cover(1999)]{cover1999elements}
Thomas~M Cover.
\newblock \emph{Elements of information theory}.
\newblock John Wiley \& Sons, 1999.

\bibitem[Crowley et~al.(2018)Crowley, Turner, Storkey, and O'Boyle]{crowley2018closer}
Elliot~J Crowley, Jack Turner, Amos Storkey, and Michael O'Boyle.
\newblock A closer look at structured pruning for neural network compression.
\newblock \emph{arXiv preprint arXiv:1810.04622}, 2018.

\bibitem[D'agostino~Sr \& Russell(2005)D'agostino~Sr and Russell]{da_2005_scree}
Ralph~B D'agostino~Sr and Heidy~K Russell.
\newblock Scree test.
\newblock \emph{Encyclopedia of biostatistics}, 7, 2005.

\bibitem[Ding et~al.(2019)Ding, Zhou, Guo, Han, Liu, et~al.]{ding2019global}
Xiaohan Ding, Xiangxin Zhou, Yuchen Guo, Jungong Han, Ji~Liu, et~al.
\newblock Global sparse momentum sgd for pruning very deep neural networks.
\newblock \emph{Advances in Neural Information Processing Systems}, 32, 2019.

\bibitem[Frankle et~al.(2020)Frankle, Dziugaite, Roy, and Carbin]{frankle2020pruning}
Jonathan Frankle, Gintare~Karolina Dziugaite, Daniel~M Roy, and Michael Carbin.
\newblock Pruning neural networks at initialization: Why are we missing the mark?
\newblock \emph{arXiv preprint arXiv:2009.08576}, 2020.

\bibitem[Giraldo et~al.(2014)Giraldo, Rao, and Principe]{giraldo_2014_measures}
Luis Gonzalo~Sanchez Giraldo, Murali Rao, and Jose~C Principe.
\newblock Measures of entropy from data using infinitely divisible kernels.
\newblock \emph{IEEE Transactions on Information Theory}, 61\penalty0 (1):\penalty0 535--548, 2014.

\bibitem[Gong et~al.(2022)Gong, Dong, Yu, and Dong]{gong2022computationally}
Tieliang Gong, Yuxin Dong, Shujian Yu, and Bo~Dong.
\newblock Computationally efficient approximations for matrix-based r{\'e}nyi's entropy.
\newblock \emph{IEEE Transactions on Signal Processing}, 70:\penalty0 6170--6184, 2022.

\bibitem[Guo et~al.(2020)Guo, Wang, Li, and Yan]{guo2020dmcp}
Shaopeng Guo, Yujie Wang, Quanquan Li, and Junjie Yan.
\newblock Dmcp: Differentiable markov channel pruning for neural networks.
\newblock In \emph{Proceedings of the IEEE/CVF conference on computer vision and pattern recognition}, pp.\  1539--1547, 2020.

\bibitem[Han et~al.(2015)Han, Mao, and Dally]{han2015deep}
Song Han, Huizi Mao, and William~J Dally.
\newblock Deep compression: Compressing deep neural networks with pruning, trained quantization and huffman coding.
\newblock \emph{arXiv preprint arXiv:1510.00149}, 2015.

\bibitem[He et~al.(2016)He, Zhang, Ren, and Sun]{he_2016_deep}
Kaiming He, Xiangyu Zhang, Shaoqing Ren, and Jian Sun.
\newblock Deep residual learning for image recognition.
\newblock In \emph{Proceedings of the IEEE conference on computer vision and pattern recognition}, pp.\  770--778, 2016.

\bibitem[He \& Xiao(2023)He and Xiao]{he2023structured}
Yang He and Lingao Xiao.
\newblock Structured pruning for deep convolutional neural networks: A survey.
\newblock \emph{IEEE transactions on pattern analysis and machine intelligence}, 2023.

\bibitem[He et~al.(2019)He, Liu, Wang, Hu, and Yang]{he2019filter}
Yang He, Ping Liu, Ziwei Wang, Zhilan Hu, and Yi~Yang.
\newblock Filter pruning via geometric median for deep convolutional neural networks acceleration.
\newblock In \emph{Proceedings of the IEEE/CVF conference on computer vision and pattern recognition}, pp.\  4340--4349, 2019.

\bibitem[He et~al.(2017)He, Zhang, and Sun]{he2017channel}
Yihui He, Xiangyu Zhang, and Jian Sun.
\newblock Channel pruning for accelerating very deep neural networks.
\newblock In \emph{Proceedings of the IEEE international conference on computer vision}, pp.\  1389--1397, 2017.

\bibitem[Howard(2017)]{howard_2017_mobilenets}
AG~Howard.
\newblock Mobilenets: Efficient convolutional neural networks for mobile vision applications.
\newblock \emph{arXiv preprint arXiv:1704.04861}, 2017.

\bibitem[Krizhevsky et~al.(2009)Krizhevsky, Hinton, et~al.]{krizhevsky_2009_learning}
Alex Krizhevsky, Geoffrey Hinton, et~al.
\newblock Learning multiple layers of features from tiny images.
\newblock 2009.

\bibitem[Krizhevsky et~al.(2012)Krizhevsky, Sutskever, and Hinton]{krizhevsky_2012_imagenet}
Alex Krizhevsky, Ilya Sutskever, and Geoffrey~E Hinton.
\newblock Imagenet classification with deep convolutional neural networks.
\newblock \emph{Advances in neural information processing systems}, 25, 2012.

\bibitem[LeCun et~al.(1998)LeCun, Bottou, Bengio, and Haffner]{lecun_1998_gradient}
Yann LeCun, L{\'e}on Bottou, Yoshua Bengio, and Patrick Haffner.
\newblock Gradient-based learning applied to document recognition.
\newblock \emph{Proceedings of the IEEE}, 86\penalty0 (11):\penalty0 2278--2324, 1998.

\bibitem[Lee et~al.(2019)Lee, Ajanthan, Gould, and Torr]{lee2019signal}
Namhoon Lee, Thalaiyasingam Ajanthan, Stephen Gould, and Philip~HS Torr.
\newblock A signal propagation perspective for pruning neural networks at initialization.
\newblock \emph{arXiv preprint arXiv:1906.06307}, 2019.

\bibitem[Li et~al.(2020)Li, Wu, Su, and Wang]{li2020eagleeye}
Bailin Li, Bowen Wu, Jiang Su, and Guangrun Wang.
\newblock Eagleeye: Fast sub-net evaluation for efficient neural network pruning.
\newblock In \emph{Computer Vision--ECCV 2020: 16th European Conference, Glasgow, UK, August 23--28, 2020, Proceedings, Part II 16}, pp.\  639--654. Springer, 2020.

\bibitem[Li et~al.(2021)Li, Liu, Yang, Peng, and Zhou]{li2021survey}
Zewen Li, Fan Liu, Wenjie Yang, Shouheng Peng, and Jun Zhou.
\newblock A survey of convolutional neural networks: analysis, applications, and prospects.
\newblock \emph{IEEE transactions on neural networks and learning systems}, 33\penalty0 (12):\penalty0 6999--7019, 2021.

\bibitem[Lin et~al.(2022)Lin, Zhang, Li, Chen, Chao, Wang, Li, Tian, and Ji]{lin20221xn}
Mingbao Lin, Yuxin Zhang, Yuchao Li, Bohong Chen, Fei Chao, Mengdi Wang, Shen Li, Yonghong Tian, and Rongrong Ji.
\newblock 1xn pattern for pruning convolutional neural networks.
\newblock \emph{IEEE Transactions on Pattern Analysis and Machine Intelligence}, 45\penalty0 (4):\penalty0 3999--4008, 2022.

\bibitem[Liu et~al.(2019)Liu, Mu, Zhang, Guo, Yang, Cheng, and Sun]{liu2019metapruning}
Zechun Liu, Haoyuan Mu, Xiangyu Zhang, Zichao Guo, Xin Yang, Kwang-Ting Cheng, and Jian Sun.
\newblock Metapruning: Meta learning for automatic neural network channel pruning.
\newblock In \emph{Proceedings of the IEEE/CVF international conference on computer vision}, pp.\  3296--3305, 2019.

\bibitem[Molchanov et~al.(2016)Molchanov, Tyree, Karras, Aila, and Kautz]{molchanov2016pruning}
Pavlo Molchanov, Stephen Tyree, Tero Karras, Timo Aila, and Jan Kautz.
\newblock Pruning convolutional neural networks for resource efficient inference.
\newblock \emph{arXiv preprint arXiv:1611.06440}, 2016.

\bibitem[Molchanov et~al.(2019)Molchanov, Mallya, Tyree, Frosio, and Kautz]{molchanov2019importance}
Pavlo Molchanov, Arun Mallya, Stephen Tyree, Iuri Frosio, and Jan Kautz.
\newblock Importance estimation for neural network pruning.
\newblock In \emph{Proceedings of the IEEE/CVF conference on computer vision and pattern recognition}, pp.\  11264--11272, 2019.

\bibitem[Niesing(1997)]{niesing_1997_simultaneous}
Jan Niesing.
\newblock Simultaneous componenet and factor analysis methods for two or more groups: a comparative study.
\newblock 1997.

\bibitem[Pelleg et~al.(2000)Pelleg, Moore, et~al.]{pelleg_2000_xmeans}
Dan Pelleg, Andrew Moore, et~al.
\newblock X-means: Extending k-means with e cient estimation of the number of clusters.
\newblock In \emph{ICML’00}, pp.\  727--734. Citeseer, 2000.

\bibitem[Phan(2021)]{huy_phan_2021_4431043}
Huy Phan.
\newblock huyvnphan/pytorch\_cifar10, January 2021.
\newblock URL \url{https://doi.org/10.5281/zenodo.4431043}.

\bibitem[R{\'e}nyi(1965)]{renyi1965foundations}
Alfr{\'e}d R{\'e}nyi.
\newblock On the foundations of information theory.
\newblock \emph{Revue de l'Institut International de Statistique}, pp.\  1--14, 1965.

\bibitem[Sadasivan et~al.(2022)Sadasivan, Malaviya, and Dasgupta]{sadasivan2022ossum}
Vinu~Sankar Sadasivan, Jayesh Malaviya, and Anirban Dasgupta.
\newblock {OSS}um: A gradient-free approach for pruning neural networks at initialization, 2022.
\newblock URL \url{https://openreview.net/forum?id=sTECq7ZjtKX}.

\bibitem[Sainath et~al.(2013)Sainath, Kingsbury, Sindhwani, Arisoy, and Ramabhadran]{sainath2013low}
Tara~N Sainath, Brian Kingsbury, Vikas Sindhwani, Ebru Arisoy, and Bhuvana Ramabhadran.
\newblock Low-rank matrix factorization for deep neural network training with high-dimensional output targets.
\newblock In \emph{2013 IEEE international conference on acoustics, speech and signal processing}, pp.\  6655--6659. IEEE, 2013.

\bibitem[Sehwag et~al.(2020)Sehwag, Wang, Mittal, and Jana]{sehwag2020hydra}
Vikash Sehwag, Shiqi Wang, Prateek Mittal, and Suman Jana.
\newblock Hydra: Pruning adversarially robust neural networks.
\newblock \emph{Advances in Neural Information Processing Systems}, 33:\penalty0 19655--19666, 2020.

\bibitem[Shneider et~al.(2023)Shneider, Rostami, Kacem, Sinha, Shabayek, and Aouada]{shneider2023impact}
Carl Shneider, Peyman Rostami, Anis Kacem, Nilotpal Sinha, Abd El~Rahman Shabayek, and Djamila Aouada.
\newblock Impact of disentanglement on pruning neural networks.
\newblock \emph{arXiv preprint arXiv:2307.09994}, 2023.

\bibitem[Simonyan \& Zisserman(2014)Simonyan and Zisserman]{simonyan_2014_very}
Karen Simonyan and Andrew Zisserman.
\newblock Very deep convolutional networks for large-scale image recognition.
\newblock \emph{arXiv preprint arXiv:1409.1556}, 2014.

\bibitem[Tanaka et~al.(2020)Tanaka, Kunin, Yamins, and Ganguli]{tanaka2020pruning}
Hidenori Tanaka, Daniel Kunin, Daniel~L Yamins, and Surya Ganguli.
\newblock Pruning neural networks without any data by iteratively conserving synaptic flow.
\newblock \emph{Advances in neural information processing systems}, 33:\penalty0 6377--6389, 2020.

\bibitem[Xu et~al.(2019)Xu, Li, Zhang, Wen, Wang, Dai, Qi, Chen, Lin, and Xiong]{xu2019trained}
Yuhui Xu, Yuxi Li, Shuai Zhang, Wei Wen, Botao Wang, Wenrui Dai, Yingyong Qi, Yiran Chen, Weiyao Lin, and Hongkai Xiong.
\newblock Trained rank pruning for efficient deep neural networks.
\newblock In \emph{2019 Fifth Workshop on Energy Efficient Machine Learning and Cognitive Computing-NeurIPS Edition (EMC2-NIPS)}, pp.\  14--17. IEEE, 2019.

\bibitem[Yang et~al.(2017)Yang, Chen, and Sze]{yang2017designing}
Tien-Ju Yang, Yu-Hsin Chen, and Vivienne Sze.
\newblock Designing energy-efficient convolutional neural networks using energy-aware pruning.
\newblock In \emph{Proceedings of the IEEE conference on computer vision and pattern recognition}, pp.\  5687--5695, 2017.

\bibitem[You et~al.(2019)You, Yan, Ye, Ma, and Wang]{you2019gate}
Zhonghui You, Kun Yan, Jinmian Ye, Meng Ma, and Ping Wang.
\newblock Gate decorator: Global filter pruning method for accelerating deep convolutional neural networks.
\newblock \emph{Advances in neural information processing systems}, 32, 2019.

\bibitem[Younesi et~al.(2024)Younesi, Ansari, Fazli, Ejlali, Shafique, and Henkel]{younesi_2024_comprehensive}
Abolfazl Younesi, Mohsen Ansari, Mohammadamin Fazli, Alireza Ejlali, Muhammad Shafique, and Jörg Henkel.
\newblock A comprehensive survey of convolutions in deep learning: Applications, challenges, and future trends.
\newblock \emph{IEEE Access}, 12:\penalty0 41180--41218, 2024.
\newblock \doi{10.1109/ACCESS.2024.3376441}.

\bibitem[Yu \& Principe(2019{\natexlab{a}})Yu and Principe]{yu_2019_Simple}
Shujian Yu and Jose~C Principe.
\newblock Simple stopping criteria for information theoretic feature selection.
\newblock \emph{Entropy}, 21\penalty0 (1):\penalty0 99, 2019{\natexlab{a}}.

\bibitem[Yu \& Principe(2019{\natexlab{b}})Yu and Principe]{yu_2019_Understanding}
Shujian Yu and Jose~C Principe.
\newblock Understanding autoencoders with information theoretic concepts.
\newblock \emph{Neural Networks}, 117:\penalty0 104--123, 2019{\natexlab{b}}.

\bibitem[Yu et~al.(2019)Yu, Giraldo, Jenssen, and Principe]{yu_2019_multivariate}
Shujian Yu, Luis Gonzalo~Sanchez Giraldo, Robert Jenssen, and Jose~C Principe.
\newblock Multivariate extension of matrix-based r{\'e}nyi's $\alpha$-order entropy functional.
\newblock \emph{IEEE transactions on pattern analysis and machine intelligence}, 42\penalty0 (11):\penalty0 2960--2966, 2019.

\bibitem[Yu et~al.(2020)Yu, Wickstr{\o}m, Jenssen, and Principe]{yu2020understanding}
Shujian Yu, Kristoffer Wickstr{\o}m, Robert Jenssen, and Jose~C Principe.
\newblock Understanding convolutional neural networks with information theory: An initial exploration.
\newblock \emph{IEEE transactions on neural networks and learning systems}, 32\penalty0 (1):\penalty0 435--442, 2020.

\bibitem[Yu et~al.(2021)Yu, Wickstrøm, Jenssen, and Príncipe]{Yu2021}
Shujian Yu, Kristoffer Wickstrøm, Robert Jenssen, and José~C. Príncipe.
\newblock Understanding convolutional neural networks with information theory: An initial exploration.
\newblock \emph{IEEE Transactions on Neural Networks and Learning Systems}, 32\penalty0 (1):\penalty0 435--442, 2021.
\newblock \doi{10.1109/TNNLS.2020.2968509}.

\bibitem[Zhang et~al.(2019)Zhang, Zhang, Chen, Sun, Ma, and Yu]{zhang2019recent}
Qianru Zhang, Meng Zhang, Tinghuan Chen, Zhifei Sun, Yuzhe Ma, and Bei Yu.
\newblock Recent advances in convolutional neural network acceleration.
\newblock \emph{Neurocomputing}, 323:\penalty0 37--51, 2019.

\bibitem[Zhao et~al.(2024)Zhao, Wang, Zhang, Han, Deveci, and Parmar]{zhao_2024_review}
Xia Zhao, Limin Wang, Yufei Zhang, Xuming Han, Muhammet Deveci, and Milan Parmar.
\newblock A review of convolutional neural networks in computer vision.
\newblock \emph{Artificial Intelligence Review}, 57\penalty0 (4):\penalty0 99, 2024.

\bibitem[Zhuang et~al.(2018)Zhuang, Tan, Zhuang, Liu, Guo, Wu, Huang, and Zhu]{zhuang2018discrimination}
Zhuangwei Zhuang, Mingkui Tan, Bohan Zhuang, Jing Liu, Yong Guo, Qingyao Wu, Junzhou Huang, and Jinhui Zhu.
\newblock Discrimination-aware channel pruning for deep neural networks.
\newblock \emph{Advances in neural information processing systems}, 31, 2018.

\end{thebibliography}
\bibliographystyle{iclr2025_conference}

\newpage
\appendix
\section{APPENDIX}


\subsection{Background}

\subsubsection{Convolutional Neural Networks}
 Convolutional Neural Networks (CNN) is a specialized type of deep neural network primarily used for processing structured grid-like data such as images  \citep{younesi_2024_comprehensive}. CNN is particularly effective in image processing tasks such as image classification or object detection, because of its ability to automatically learn and extract \emph{hierarchical features} from the input data. Different CNN architectures have been introduced for image processing tasks, including LeNet \citep{lecun_1998_gradient}, AlexNet \citep{krizhevsky_2012_imagenet}, Visual Geometry Group (VGG) \citep{simonyan_2014_very}, Residual Network (ResNet) \citep{he_2016_deep} and MobileNet \citep{howard_2017_mobilenets}. 

\begin{figure}[h]
\begin{center}
\includegraphics[width=0.8\textwidth]{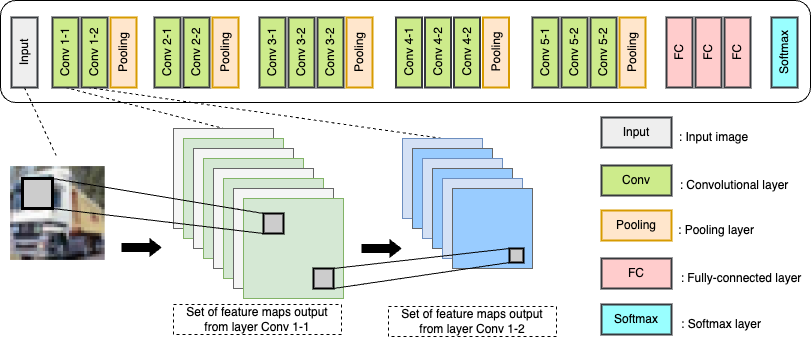}
\end{center}
\caption{Illustration of the process of a sample CNN model.}
\label{fig:process_of_CNN}
\end{figure}

A CNN architecture generally consists of an input layer, a stack of alternating convolutional and pooling layers, several fully connected layers, and an output layer at the end \citep{zhao_2024_review}. The top panel in Fig. \ref{fig:process_of_CNN} shows the VGG-16 architecture, which includes 13 convolutional layers and 3 fully connected layers. Each convolutional layer contains a set of filters. A convolution operation involves sliding a filter over the input image, multiplying the filter values by the pixel values at corresponding positions in the input image, and summing the results to obtain a feature map. By applying various filters to the input image, a set of feature maps is generated, as shown in Fig. \ref{fig:process_of_CNN}. When multiple convolutional layers are stacked, the later layers capture more representative features of the input image. We will use the VGG-16 architecture as the main example for implementation in this paper, but all the discussion and developed algorithms can be applied to any CNN structure.

\subsubsection{Multivariate mutual information using Rényi entropy} 

Our proposed CNN pruning method is based on computing the conditional mutual information between the features extracted in the same layer and in different layers of the CNN. Each feature is treated as a multivariate random variable in matrix form. The test data after being processed through the trained CNN provides samples or realizations of each random feature at each layer. Next, we discuss the method used for computing the mutual information (MI) and conditional mutual information (CMI) subsequently.

\textbf{Rényi Entropy and Mutual Information Computation:} 
To estimate MI between random variables, we rely on the Rényi's $\alpha$-order entropy $H_\alpha(X)$ ~\citep{renyi1965foundations}, defined as 
\begin{equation}
    H_\alpha(X) = \frac{1}{1 - \alpha} \log \left( \int_X p^\alpha(x) \, dx \right),
\end{equation}
where $X$ is a continuous random variable with the probability density function (PDF) $p(x)$, and $\alpha$ is a positive constant.  
Rényi entropy extends the well-known Shannon entropy which is obtained when the parameter $\alpha$ approaches $1$~\citep{renyi1965foundations}.

Calculating Rényi entropy requires knowing the PDF, which limits its application in data-driven context. To overcome this, we employ a matrix-based $\alpha$-order Rényi entropy calculation \citep{giraldo_2014_measures} 
which computes Rényi's $\alpha$-order entropy using the eigenspectrum of a normalized Hermitian matrix, derived by projecting data into a Reproducing Kernel Hilbert Space (RKHS) \citep{gong2022computationally}:
\begin{equation}
S_\alpha(G) = \frac{1}{1 - \alpha} \log_2 \left( \text{tr}(G^\alpha) \right) = \frac{1}{1 - \alpha} \log_2 \left( \sum_{i=1}^n \lambda_i^\alpha(G) \right),    
\label{eq:Renyi_matrix_based}
\end{equation}
where $G$ is a normalized kernel matrix obtained from the data and $\lambda_i(G)$ are the eigenvalues of $G$.

For a given CNN, to construct matrix $G$, we first extract latent features from the CNN by feed-forwarding the training data to each CNN layer. This process provides for each layer a feature matrix $\mathbf{X}^{N \times d}$, where each row represents a $d$-dimensional feature vector of a data sample. We then compute the kernel matrix $\hat{G}$ from these features using a kernel function $\varphi(x_i, x_j)$ that measures the similarity between feature vectors $x_i$ and $x_j$. In our experiment, we use the RBF kernel $\varphi(x_i, x_j)=\text{exp}(-{||x_i-x_j||^2}/{(2\sigma^2)})$. Next, we normalize the kernel matrix $\hat{G}$ to obtain the normalized kernel matrix $G$. The normalization ensures $G$ is symmetric and its eigenvalues are within the range $[0, 1]$. 



For multiple variables, the matrix-based Rényi's $\alpha$-order  joint entropy of $L$ variables is computed as \citep{yu_2019_multivariate}
\begin{equation} \label{eq:mi_multivar}
S_\alpha(G_1, G_2, \ldots, G_L) = S_\alpha \left( \frac{G_1 \circ G_2 \circ \cdots \circ G_L}{\text{tr}(G_1 \circ G_2 \circ \cdots \circ G_L)} \right),
\end{equation}
where $(G^k)_{ij} = \varphi_k(x_i^k, x_j^k)$, with $k \in \{1,...,L\}$ denotes the normalized kernel matrix of the k$^{th}$ variable, and $\varphi_k$: $\mathcal{X}^k \times \mathcal{X}^k \mapsto \mathbb{R}$ is the k$^{th}$ positive definite kernel, and $\circ$ denotes the Hadamard product.

Using Rényi entropy, the matrix-based Rényi's $\alpha$-order mutual information $I_\alpha(\cdot; \cdot)$ is computed as
\begin{equation}
    I_\alpha(G; G_1, \ldots, G_L) = S_\alpha(G) + S_\alpha(G_1, \ldots, G_L) - S_\alpha(G_1, \ldots, G_L, G)
\end{equation}





\begin{algorithm}[t]
\caption{CMI permutation test \citep{yu_2019_Simple}}
\label{alg:cmi_p_test}
\begin{algorithmic}[1]
\STATE \textbf{Input:} Selected ordered set of feature maps $F_k^{s}$, remaining feature maps $F_k^r$, class labels $Y$, selected feature map $f$ (in $F_k^r$), permutation number $P$, significance level $\alpha$


\STATE \textbf{Compute:} Estimate $I(\{F_k^r - f\}; Y \mid \{F_k^s, f\})$ 

\FOR{$i = 1$ to $P$}
    \STATE Randomly permute $f$ to obtain $\tilde{f}_i$
    \STATE Estimate $I(\{F_k^r - \tilde{f}_i\}; Y \mid \{F_k^s, \tilde{f}_i\})$
\ENDFOR 
\STATE \textit{Evaluate the significance:} 
\IF{$\frac{1}{P} \sum_{i=1}^{P} \mathbf{1}[I(\{F_k^r - f\}; Y \mid \{F_k^s, f\}) \geq I(\{F_k^r - \tilde{f}_i\}; Y \mid \{F_k^s, \tilde{f}_i\})] \leq \alpha$}
    \STATE $F_k^s$ $\leftarrow$ $F_k^s \cup f$
    \STATE \texttt{decision} $\leftarrow$ Continue feature map selection
\ELSE
    \STATE \texttt{decision} $\leftarrow$ Stop feature map selection
    \STATE $N \leftarrow |F_k^s|$
\ENDIF

\STATE \textbf{return} \texttt{decision}, $N$ 
\end{algorithmic}
\end{algorithm}

\textbf{Conditional Mutual Information Computation using Rényi Entropy:} Conditional mutual information (CMI) quantifies the amount of information shared between two random variables, $X$ and $Y$, given the knowledge of a third variable $Z$. Typically, it is expressed using Shannon entropy as
\begin{equation}
    I(X;Y|Z) = H(X,Z) + H(Y,Z) - H(X,Y,Z) - H(Z)
\end{equation}
Using Rényi entropy, CMI can be generalized as the matrix-based Rényi $\alpha$-order CMI:
\begin{equation}\label{eq:cmi_matrix}
    I_\alpha(G_X;G_Y|G_Z) = S_\alpha(G_X, G_Z) + S_\alpha(G_Y, G_Z) - S_\alpha(G_X, G_Y, G_Z) - S_\alpha(G_Z),
\end{equation}
where $G_X, G_Y, G_Z$ are the normalized kernel matrices defined on the data samples of the variables $X$, $Y$, and $Z$, respectively.

\subsection{Permutation Test}
We describe in this section the \emph{Permutation Test} used by \citep{yu_2019_Simple} to quantify the impact of a new feature map $f$ on the model accuracy. Specifically, for a new feature $f$, CMI permutation test creates a random permutation $\tilde{f}$ from $\{f \cup F^{s}_k$\}, and computes the new CMI value between the output $Y$ and the set of unselected features, conditioned on the permutation set $\tilde{f}$. The algorithm then compares this new CMI value with the original CMI that is conditioned on the original set $\{f \cup F^{s}_k$\} to determine whether the contribution of feature $f$ on the output is significant. 
Specifically, if the CMI value of the permutated feature set is not significantly smaller than the original CMI value, the permutation test will discard feature $f$, as $f$ does not capture the spatial structure in the input data, and stop the feature selection process. However, applying CMI permutation method on CNN models leads to the retention of very few filters \citep{Yu2021}, resulting in a significant drop in the model accuracy. We describe the CMI permutation test as used for feature selection in \citep{Yu2021} in Algorithm \ref{alg:cmi_p_test}.


\subsection{Different Cutoff Point Approaches on Per-Layer CMI}

In this section, we compare three approaches, Permutation test, Scree test and X-means, for determining the cutoff point of CMI values and evaluate their effectiveness on \emph{per-layer CMI}. {Here we prune each layer individually without pruning any other layers, and evaluate the accuracy performance of the resulting pruned model with one layer pruned.} 
The results are provided in Table \ref{tab:per-layer_cmi_diff-cutoff}, showing that the Permutation test retains high accuracy in only $4$ out of $13$ convolutional layers, while both the Scree test and X-means maintain high accuracy in all layers. The impact of using the Permutation test to prune all layers is even more dramatic as seen by the results in Table \ref{tab:fm-pruning-comparison}.

\begin{table}[t]
\caption{Comparison of Permutation Test, Scree Test, and X-Means on \emph{Individual Layer pruning} with per-layer CMI. {Each test accuracy value is shown for the pruned model obtained by pruning \emph{only} the current layer.} 
Accuracy values above 90\% are in \textbf{bold}.}
\label{tab:per-layer_cmi_diff-cutoff}
\begin{center}
\begin{tabular}{cccccccc}
\bf  & \bf  & \multicolumn{2}{c}{\bf PERMUTATION TEST} & \multicolumn{2}{c}{\bf SCREE TEST} & \multicolumn{2}{c}{\bf X-MEANS} \\
\hline
\bf Layer & \bf Total & \bf \#Filters & \bf  & \bf \#Filters & \bf  & \bf \#Filters & \bf \\
\bf No. & \bf \#Filters & \bf Selected & \bf Acc. & \bf Selected & \bf Acc. & \bf Selected & \bf Acc. \\
\hline
\\
{1} & 64 & {2} & 12.83\% & 49 & \bf 94.00\% & 47 & \bf 94.00\% \\
{2} & 64 & {2} & 9.99\% & 60 & \bf 92.89\% & 47 & \bf 91.27\% \\
{3} & 128 & {2} & 10.00\% & 124 & \bf 93.40\% & 111 & \bf 93.16\% \\
{4} & 256 & {8} & 8.40\% & 109 & \bf 91.91\% & 111 & \bf 92.39\% \\
{5} & 256 & {2} & 9.99\% & 229 & \bf 93.17\% & 223 & \bf 92.45\% \\
{6} & 256 & {1} & 9.99\% & 247 & \bf 93.44\% & 239 & \bf 92.48\% \\
{7} & 512 & {19} & 20.95\% & 238 & \bf 93.71\% & 159 & \bf 91.71\% \\
{8} & 512 & {17} & 10.23\% & 414 & \bf 93.68\% & 265 & \bf 92.58\% \\
{9} & 512 & {23} & 80.63\% & 218 & \bf 93.13\% & 244 & \bf 93.58\% \\
{10} & 512 & {19} & \bf 93.97\% & 192 & \bf 93.71\% & 140 & \bf 93.62\% \\
{11} & 512 & {19} & \bf 94.00\% & 215 & \bf 93.66\% & 195 & \bf 93.59\% \\
{12} & 512 & {79} & \bf 94.00\% & 326 & \bf 94.02\% & 136 & \bf 93.79\% \\
{13} & 512 & {359} & \bf 93.78\% & 448 & \bf 93.92\% & 51 & \bf 93.53\% \\
\end{tabular}
\end{center}
\end{table}


\subsection{Full CMI versus Compact CMI on Forward Pruning}

In this section, we present the experimental results of Forward Pruning in \ref{alg:forward_cmi} with two methods for ranking features and computing CMI values: \emph{Full CMI} (Section \ref{sec:full_cmi}) and \emph{Compact CMI} (Section \ref{sec:compact_cmi}), using Scree test as the cutoff point method. Table \ref{tab:forward_full_vs_compact} presents the results of the number of selected filters and the corresponding accuracy of the pruned model after iteratively pruning each layer. We observe that, for the first 12 layers, Full CMI retains more filters than Compact CMI and hence results in a smaller decrease in accuracy. However, in the last CNN layer, Full CMI retains very few filters, leading to the significant drop in the pruned model's accuracy. On the other hand, Compact CMI has a higher pruned percentage by retaining fewer filters in most layers (except the last one) while maintaining relatively consistent accuracy throughout all layers. 

\begin{table}[t]
\caption{Full CMI versus Compact CMI on Forward Pruning with Scree test, using \emph{Zero weight} pruning where the non-selected filters are set to $0$ but not removed from the CNN. {Each test accuracy value is shown for the pruned model obtained by pruning all layers from the first layer up to and including the current layer, without retraining.}}
\label{tab:forward_full_vs_compact}
\begin{center}
\begin{tabular}{cccccc}
\bf  & \bf  & \multicolumn{2}{c}{\bf FULL CMI} & \multicolumn{2}{c}{\bf COMPACT CMI} \\
\bf Layer & \bf Total & \bf \#Filters & \bf  & \bf \#Filters & \bf  \\
\bf No. & \bf \#Filters & \bf Selected & \bf Acc. & \bf Selected & \bf Acc. \\
\hline
\\
{1} & 64 & {49} & 94.00\% & 49 & 94.00\% \\
{2} & 64 & {59} & 93.55\% & 59 & 93.59\% \\
{3} & 128 & {124} & 93.48\% & 108 & 92.95\% \\
{4} & 256 & {125} & 93.47\% & 125 & 92.95\% \\
{5} & 256 & {252} & 93.26\% & 209 & 91.37\% \\
{6} & 256 & {252} & 93.04\% & 251 & 91.33\% \\
{7} & 512 & {248} & 92.95\% & 248 & 91.24\% \\
{8} & 512 & {504} & 92.93\% & 355 & 90.19\% \\
{9} & 512 & {505} & 92.93\% & 405 & 89.81\% \\
{10} & 512 & {501} & 92.95\% & 197 & 88.73\% \\
{11} & 512 & {507} & 92.95\% & 323 & 87.71\% \\
{12} & 512 & {505} & 92.95\% & 255 & 88.19\% \\
{13} & 512 & {11} & 37.79\% & 408 & 87.38\% \\
\end{tabular}
\end{center}
\end{table}

\subsection{Comparison Between Features Retained by Scree test and X-means}

To examine in more detail the difference between Scree test and X-means, we analyze the selected feature sets of each approach {using Bi-directional pruning with Compact CMI computation}. 
Table \ref{tab:cmp_scree_xmeans} shows the comparison. 
The \textit{Overlap} presents the percentage of feature maps that are retained by both Scree test and X-means, relative to the total number of feature maps in a given layer. This "Overlap" measure provides insight into the agreement between the two cutoff point approaches regarding which feature maps are essential. \textit{Scree test Only} and \textit{X-means Only} represent the percentage of feature maps retained exclusively by the Scree test and X-means, respectively, relative to the total number of features retained by each approach. 
{We can see that the overlap of selected features between the two approaches is highest for Layer 6 and gradually decreases the farther away from this layer. This overlap percentage is in agreement with the percentage of filters pruned shown for each approach, as Layer 6 has the lowest percentage pruned for both methods.  We note also that the starting layer for pruning with Scree-test is Layer 10, and with X-means is Layer 13. The percentage of filters pruned is highest for each method at its starting layer and decreases from there, but not necessarily in a strictly decreasing order the farther away from the starting layer. This result is quite curious and shows that different sets of filters can be pruned at each layer depending on the cutoff point method while still preserving the final accuracy within a relatively reasonable range. The final re-trained pruned model obtained with either Scree-test or X-means has a test accuracy within $1.01\%$ of the original unpruned model (as shown in Table \ref{tab:fm-pruning-comparison}).}

\begin{table}[t]
\caption{Comparison of Shared and Exclusive retained feature maps between Scree test and X-means on Bi-directional pruning with Compact CMI. {The "Overlap" column shows the percentage of overlapping selected filters, and the last two columns show the individual percentage of filters pruned, all relative to the total number of filters in each layer. The "Only" columns show the percentage of uniquely selected filters relative to the total number of selected filters in each method. The star ($^\star$) indicates the starting layer for pruning in each method.} }
\label{tab:cmp_scree_xmeans}
\begin{center}
\begin{tabular}{ccccccc}
\bf LAYER & \bf OVERLAP  & \bf SCREE TEST & \bf X-MEANS & \multicolumn{2}{c}{\bf \%FILTERS PRUNE} \\
\bf Index & \bf & \bf Only & \bf Only & \bf Scree Test & \bf X-Means \\
\hline
\\
1 & 68.75\% & \bf 0.00\% & 6.38\% & 31.25\% & 26.56\% \\
2 & 73.44\% & 22.95\% & \bf 0.00\% & 4.69\% & 26.56\% \\
3 & 86.72\% & 10.48\% & \bf 0.00\% & 3.13\% & 13.28\% \\
4 & 86.72\% & 8.26\% & \bf 0.00\% & 5.47\% & 13.28\% \\
5 & 92.19\% & \bf 0.00\% & 1.26\% & 7.81\% & 6.64\% \\
6 & 93.36\% & 4.78\% & \bf 0.00\% & 1.95\% & 6.64\% \\
7 & 83.20\% & 0.47\% & 4.48\% & 16.41\% & 12.89\% \\
8 & 55.86\% & 30.07\% & \bf 0.00\% & 20.12\% & 44.14\% \\
9 & 52.15\% & \bf 0.00\% & 44.49\% & 47.85\% & 6.05\% \\
10 & 26.17\% & 30.21\% & 4.29\% & \bf 62.50 \% ($ ^\star$) & 72.66\% \\
11 & 27.73\% & 29.35\% & 15.98\% & 60.74\% & 66.99\% \\
12 & 47.46\% & 2.80\% & 26.81\% & 51.17\% & 35.16\% \\
13 & 9.96\% & 85.51\% & \bf 0.00\% & 31.25\% & \bf 90.04\%  ($ ^\star$) \\
\end{tabular}
\end{center}
\end{table}


\subsection{Analysis on Pruning Types: Zero Weights Versus Actual Pruning}

In this experiment, we consider two types of pruning: \textit{Zero weight}, which sets the pruned weights to zero while keeping the network structure unchanged, and \textit{Actual pruning}, which completely removes the pruned weights from the network, thereby reducing the number of parameters and memory usage. During \textit{Actual pruning}, as we focus on CNN layers, we leave the last CNN layer unpruned to preserve its connections to the following fully connected layer. 

{These two pruning types also involve a difference in the BatchNorm layer operation following each pruned CNN layer. In Zero-weight pruning, we set the pruned filters to zero without adjusting the BatchNorm layer. In actual pruning, however, the pruned filters are completely removed from the CNN model, hence the shape of each pruned CNN layer changes and we adjust the BatchNorm operation accordingly to match the smaller shape. These adjustments lead to different test accuracies between Zero-weight and Actual pruning for the pruned models.}

Table \ref{tab:cnn-ordering-2-pruning-types} shows the comparison between Zero-weight and Actual pruning with different CNN pruning and CMI computation methods. We use the Scree test for selecting the cutoff point. The results show that \textit{Zero-weight} pruning leads to higher {pruned percentage} compared to \textit{Actual pruning} for three out of the four settings. However, \emph{Actual pruning} consistently leads to higher test accuracy for the final pruned model across all settings. We also note that Bi-directional pruning with compact CMI achieves the best performance, with highest {pruned percentage} in both pruning types while still maintaining high accuracy even before re-training. 

\begin{table}[t]
\caption{\textcolor{black}{Zero weight versus Actual pruning using Scree test Cutoff Point with various CMI Computation Approaches and Pruning Directions}}
\label{tab:cnn-ordering-2-pruning-types}
\begin{center}
\begin{tabular}{llcc}
\bf CNN PRUNING & \bf FEATURES ORDERING & \multicolumn{2}{c}{\bf {PRUNING TYPE}} \\
&  & \bf Zero-weight & \bf Actual pruning \\
\hline 
\\
\multicolumn{2}{l}{\bf Filters Pruned Percentage} & & \\
Forward pruning & full CMI & 13.78\% & 2.18\% \\
Forward pruning & compact CMI  & 29.17\% & 26.70\% \\
Bi-directional pruning & full CMI  & 34.04\% & 30.12\% \\
Bi-directional pruning & compact CMI  & \bf 35.56\% & \bf 36.15\% \\
\hline 
\\
\multicolumn{3}{l}{\bf Parameters Retained (unpruned model: 33.647 M)} & \\
Forward CMI & full CMI  & - & 33.196 M \\
Forward CMI & compact CMI  & - & 25.7 M \\
Bi-directional pruning & full CMI  & - & 25.643 M \\
Bi-directional pruning & compact CMI  & - & \bf 24.618 M \\
\hline 
\\
\multicolumn{2}{l}{\textbf{Accuracy \textit{before Retraining} (unpruned model: 94.00\%)}} & & \\
Forward CMI & full CMI  & 37.79\% & 93.02\% \\
Forward CMI & compact CMI  & 87.38\% & 90.17\% \\
Bi-directional pruning & full CMI  & 84.95\% & 88.59\% \\
Bi-directional pruning & compact CMI  & {\bf 82.12\%} & {\bf 90.95\%} \\
\hline
\\
\multicolumn{4}{l}{\bf Accuracy after Retraining} \\
Forward CMI & full CMI  & - & 93.67\% \\
Forward CMI & compact CMI  & - & 93.33\% \\
Bi-directional pruning & full CMI  & - & 93.25\% \\
Bi-directional pruning & compact CMI  & - & \bf 93.68\% \\

\end{tabular}
\end{center}
\end{table}

Finally, Table \ref{tab:cutoff-point-2-pruning-types} shows the comparison between \textit{Zero-weight} and \textit{Actual pruning} using different cutoff point methods. The CNN pruning and CMI computation methods are Bi-directional pruning and Compact CMI, respectively. The results show that the \textit{pruned percentage} of Permutation test is highest compared to other cutoff point methods in both pruning types. However, Permutation test results in extremely low accuracy both before and after retraining, making it unsuitable for practical purposes. The Scree test provides highest accuracy among all methods in both pruning types.

\begin{table}[!h]
\caption{\textcolor{black}{Zero weight vs. Actual pruning on Bi-directional Pruning with Compact CMI using Various Cutoff Point Approaches}}
\label{tab:cutoff-point-2-pruning-types}
\begin{center}
\begin{tabular}{llcc}
\multicolumn{2}{c}{\bf {CUTOFF POINT METHOD}} & \multicolumn{2}{c}{\bf {PRUNING TYPE}} \\
&  & \bf Zero-weight & \bf Actual pruning \\
\hline 
\\
\multicolumn{2}{l}{\bf Filters Pruned Percentage} & & \\
Permutation test &  & \bf 75.50\% & \bf 81.79\%  \\
Scree test &   & 35.56\% & 31.77\% \\
X-mean &  & 41.38\% & 34.67\%  \\
\hline 
\\
\multicolumn{4}{l}{\bf Parameters Retained (unpruned model: 33.647 M)} \\
Permutation test & & - & \bf 19.379 M \\
Scree test & & - & 24.618 M \\
X-means & & - & 25.01 M \\
\hline 
\\
\multicolumn{4}{l}{\bf Accuracy before Retraining (unpruned model: 94.00\%)} \\
Permutation test & & 9.99\% & 9.99\% \\
Scree test &  & \bf 82.12\% & \bf 90.95\% \\
X-means &  & 22.09\% & 83.56\% \\
\hline
\\
\multicolumn{4}{l}{\bf Accuracy after Retraining} \\
Permutation test &   & - & 10.02\% \\
Scree test &   & - & \bf 93.68\% \\
X-means &   & - & 92.99\% \\

\end{tabular}
\end{center}


\end{table}

\end{document}